\documentclass{article}

\usepackage{PRIMEarxiv}

\usepackage{fancyhdr}
\usepackage[utf8]{inputenc} 
\usepackage[T1]{fontenc}    
\usepackage{hyperref}       
\usepackage{url}            
\usepackage{booktabs}       
\usepackage{amsfonts}       
\usepackage{nicefrac}       
\usepackage{microtype}      
\usepackage{xcolor}         

\usepackage{graphicx,subcaption,lipsum}
\usepackage{algorithm}
\usepackage{algpseudocode}
\usepackage{verbatim}       

\usepackage{amsmath,amssymb,amsfonts}
\usepackage{amsthm}
\newtheorem{theorem}{Theorem}[section]

\usepackage{algorithm}
\usepackage{algpseudocode}
\title{A new methodology to decompose a parametric domain using reduced order data manifold
 in machine learning
}
\author{
  Chetra Mang, Axel TahmasebiMoradi, Mouadh Yagoubi \\
  IRT SystemX \\
  2 Bd Thomas Gobert, 91120 Palaiseau, France \\
  \texttt{\{chetra.mang, a.tahmasebimoradi, mouadh.yagoubi\}@irt-systemx.fr} \\
}

\begin{document}
\maketitle

\begin{abstract}
  We propose a new methodology for parametric domain decomposition using iterative principal component analysis. Starting with iterative principle component analysis, the high dimension manifold is reduced to the lower dimension manifold. Moreover, two approaches are developed to reconstruct the inverse projector to project from the lower data component to the original one. Afterward, we provide a detailed strategy to decompose the parametric domain based on the low dimension manifold. Finally, numerical examples of harmonic transport problem are given to illustrate the efficiency and effectiveness of the proposed method comparing to the classical meta-models such as neural networks.
\end{abstract}

\keywords{Domain decomposition \and Dimension reduction \and Latent space \and Machine learning \and Manifold}

\section{Introduction}

The mathematical modeling of variables and the mechanisms by which they interact to explain the observations (dataset) is the foundation of machine learning algorithms. Complex datasets, like speech, brain signals, or natural images, typically require advanced machine learning models in order to accurately represent the probability distribution of the data. With dozens to hundreds of layers, feed-forward deep neural networks—which produce a very large set of model parameters—are the foundation for the majority of complex machine learning models. 
The underlying premise of standard machine learning training algorithms is that datasets are large enough to pursue the training of very large models with success. However, large datasets are either too expensive or not always available. 

Data augmentation, or artificially generating additional samples by applying a composition of random class-preserving transformations on available data samples, is a common practice, especially when there is a limited amount of data. Furthermore, current machine learning algorithms were created with the assumption of perfect input data samples in which the data is well distributed in the parametric space.

One way to tackle the problem of not having a perfect dataset is domain decomposition. Since global models can be quite limiting when several kinds of physical regimes are present, the goal of the parameter space decomposition is to combine various local models in the parameter space into a final one in order to improve the overall accuracy. Local models are more reliable and specialized, but they are limited to a smaller portion of the parameter space. This approach is particularly useful when the dimension of input data is large and there is a limited number of samples. In the 1990s, a technique known as mixture of experts (one of the earliest domain decomposition technique) was developed to address issues with regression and classification \cite{jordan1994hierarchical}. In what follows, we are going to present a few notable works regarding the domain decomposition technique in the next section.

\subsection{Related work}
A domain decomposition methodology served as the foundation for quick manifold learning algorithm was introduced in \cite{zhang2005domain}. The authors developed the interface problem solution that can merge the embedding on the two subdomains into an embedding on the entire domain, starting with the set of sample points divided into two subdomains. They used matrix perturbation theory to give a thorough analysis of the errors caused by the merging process. Numerical examples were provided to demonstrate the effectiveness and efficiency of the suggested techniques. 

An approach to speed up Reduced Order Models (ROMs) was proposed by \cite{cizmas2008acceleration}. The dataset were composed of transient and steady state solutions; and thus, the dataset was divided into two parts. Using a predetermined threshold that was specified, the algorithm correctly identified the transition between the two regimes. Since every subset was linked to a particular regime, the initial modes had more energy and, consequently, required fewer modes to reconstruct the ROM, which resulted in a roughly 50\% boost in performance.

In 2010, a domain decomposition strategy to balance the learning times and increase global accuracy was proposed \cite{sauget2010efficient}. Its principle is based on a domain decomposition on the input parameters of the learning data set while accounting for its complexity. The algorithm was experimentally evaluated both qualitatively and quantitatively using real data sets. They validated that their algorithm behaved well in terms of both quality and performance. The minor variations in the subdomain learning times demonstrated how well the proposed method performs in a practical setting. 


A recursive subdivision of the parametric domain based on quadtree was introduced by \cite{haasdonk2011training}. Every element in the grid was refined into many subdomains if the error bound or basis size exceeded specified criteria. Until all of the thresholds were not crossed, the process was then repeated. Two-dimensional cases were used to evaluate the approach, and the results showed a significant reduction in prediction errors. Nevertheless, the sensitivity of this quadtree-based approach depends greatly on the size of the parameter field.


To address the dimensionality problem, a domain-decomposed approach that simultaneously decompose the spatial domain and the parameter domain was presented in  \cite{xu2024domain}.  A local Karhunen-Loeve (KL) expansion was built on each subdomain, and more crucially, a local inversion problem was independently solved in parallel in a lower-dimensional space. Following the generation of local posterior samples by means of subdomain Markov chain Monte Carlo (MCMC) simulations, a new projection process was proposed in order to efficiently reconstruct the global field. In addition, an adaptive Gaussian process-based fitting strategy was applied to the domain decomposition interface conditions. 

In the context of physical problems, the parameter space decomposition can be viewed as a remedy to the curse of dimensionality problem. This is because local models that are better suited to physics require less sophisticated model than global approaches. However, it is not an easy task to pre-determine the different regimes in a complex physical problem. The aforementioned difficulty is addressed the proposed methodology in this paper.

\subsection{Main contribution}
The main contributions of this work are: (1) proposing an iterative principle component analysis (iPCA) to reduce the dimension of the data manifold, and subsequently an inverse model for low dimensional data to original data, (2) introducing a methodology to construct a stretched 1-manifold of data that is used with the decomposition technique, and (3) proposing a simple yet effective technique to decomposed the stretched 1-manifold according to a user-defined criterion.

The rest of the paper is structured as follows: In
Section 2, we define the methodology which includes the iPCA and its inverse projector, connected curve of 1-manifold and its stretched manifold, a decomposition algorithm based on line similarity segment for the parametric domain decomposition, and strategies for interpolating in the stretched manifold latent space. In section 3, we present harmonic transport problem which is used to generate data for the numerical experiments. In addition, we present results corresponding to the generated dataset. Afterwards, we investigate the methodology and analyze the results in. We conclude and summary our work in section 4.

\section{METHODOLOGY}
\label{methodo}

This section describes the methodology used for parametric decomposition carried out in low dimensional manifold for data manifold learning. 

\subsection{Reduced dimension component using iterative principal component analysis}

Consider a parametric domain $X \subset \mathbb{R}^{n\times p}$ where $n$ and $p$ are size and dimension of $X$, respectively. First, the data $X$ is needed to be standardized to get the zero mean and the variance equal to 1. Then, the singular value decomposition (SVD) is performed on data $X=USV^T$ where $U$, $V$ and $S$ are left and right singular vectors, and square diagonal matrix, respectively. After that, some column vectors of either $U$ or $V$ are kept to obtain the satisfactory reduced number of components. However, the higher the number components are reduced, the more the data information is lost. This information is derived from each component variance according to the total data variance. It is called explained variance ratio (EVR) \cite{holland2008principal}. To decrease the information loss, the number of data component coordinates is to be reduced by keeping as large EVR as possible. To reach the target number of data component coordinates, one needs to reduce further the number of the coordinates such that the EVR is above target $r\%$ or reduce the small number of components $s$ of either the vector columns of $U$ or $V$. Without loss of generality, we develop the theorem based on the reduced number of column vectors of $V$. This procedure is performed iteratively till the target number of data component coordinates is reached.

\begin{theorem}\label{thm:ipca}
Given the original data $X$ of dimension $p\in \mathbb N$, there exists $k\in \mathbb N$ such that a reduced components $X_k$ of dimension $p_k\in \mathbb N$ and $p_k<p$ is expressed as $X_k=\pi_k(X)=X\pi_k$ where $\pi_k = \prod\limits_{j\in \{1...k\}}V_j$ and $V_j$ is the singular vector of the SVD of $X_{j-1}$ by keeping as much information as possible at each iteration $j$. \
\end{theorem}

\subsubsection{First approach - pseudo-inverse}

As stated in theorem~\ref{thm:ipca}, at each iteration of SVD, the maximum information while iteratively reducing the data components is kept. At iteration $j$, $X_j=X_{j-1}V_j$, $V_j$ has the maximum singular value as possible. However, $V_jV_j^T$ is the singular matrix. To reverse the projection, we need to inverse the matrix $V_jV_j^T$. It is only possible by using pseudo inverse introduced by Moore and Penrose \cite{barata2012moore}. Therefore, $X_{j-1}=X_jV_j^T(V_jV_j^T)^{+}$ where $A^{+}$ is pseudo inverse of $A$. Let note $F_j=V_j^T(V_jV_j^T)^{+}$. 

\begin{theorem}\label{thm:pinv}
Let $X$ be the original data and for a given $k\in \mathbb N$, $X_k=X\pi_k$ which are defined in \ref{thm:ipca}. For any $X'_k=\pi_k(X')$, then $X'=\pi_k^{-1}(X'_k)=X'_k\pi_k^{-1}$ where $\pi_k^{-1}=\prod\limits_{j\in \{1...k\}} F_{k-j+1}$.
\end{theorem}

\subsubsection{Second approach - component complement inverse}

The first approach is relied on the pseudo inverse of the singular matrix of each iteration. Even if at each iteration, we keep the maximum information, the accuracy of the inverse projection could still questionable. We propose the second approach based on the information of the component complement. 

Consider at iteration $j$, we reduce the singular vector $V_{j-1}$ to $V_j$, then the eliminated column vectors of $V_{j-1}$ is $V_j^c$ which is defined as \textbf{singular vector complement} of $V_j$ and $X_j^c=X_{j-1}V_j^c$ is defined as \textbf{component complement} of $X_j$.

\begin{theorem}\label{thm:ccinv}
Let $X$ and $X'$ be the training and test data, respectively. For a given $k\in \mathbb N$, $X_k=X\pi_k$ which is defined in \ref{thm:ipca}, then for a given $X_k'$ of a reduce data component coordinate, its inverse in original data manifold $X'$ is the solution to the following system of equations :
\[
\left\{
    \begin{array}{l}
        X'_{j-1}=X'_jV^T_j + X^{c'}_jV^{cT}_j\\
        \min\limits_{X^{c'}_j}d_E((X'_j,X^{c'}_j), (X_j,X^c_j))\\
    \end{array}
\right.
\]
where $X_j,X^c_j,V_j,V^c_j$ are known, $j\in\{1...k\}$ and $d_E$ is a Euclidean distance.
\end{theorem}

\subsection{Stretched reduced order data manifold}

This section details a strategy for constructing a stretched reduced order data manifold.
Consider the dataset $(X,Y)$ where $X$ is the parametric domain of size and dimension $n$ and $p$, respectively, and $Y$ is the output domain of size and dimension $n$ and $q$, respectively. The dataset forms a data manifold. We suppose that $X$ is an open set and the data manifold is a continuous multivariable vectorial function.

\subsubsection{1-manifold data}

We apply the iterative principal component approach to reduce the standardized parametric and output data $(X,Y)$ by building the corresponding projectors $\pi_x$ and $\pi_y$ as defined in \ref{thm:ipca} to reduce the dimension $p$ and $q$ of $X$ and $Y$ respectively to one dimension. Thus, the 1-manifold $(\tilde X,\tilde Y)$ is obtained. It is worth noting that $(\tilde X,\tilde Y)$ are in fact $(X_k,Y_k)$ that are reduced into 1 dimension. 
As the original manifold is assumed to be continuous, then the 1-manifold is also assumed to be continuous as well. As the original manifold is a function defined on an open set, then the 1-manifold is an open curve (disjoint extremities). 
To obtain such 1-manifold, ball pivoting algorithm (BPA) \cite{bernardini1999ball} is used to triangulate points cloud of $(\tilde X,\tilde Y)$. By doing so, we get a 2D-triangulated domain. Providing the connectivity of $(\tilde X,\tilde Y)$, we simply connect the edges of the triangulated points cloud from bottom to top to construct the 1-manifold. 

\subsubsection{1-manifold stretching}

The original manifold by assumption is a continuous function. When we reduce its dimension to one, the manifold is folded into a continuous complex curve according to its information loss. However, this continuous complex curve (a.k.a 1-manifold) may not be a function. This section introduces the concept of transforming a complex curve of 1-manifold into a 1D scalar function. 

\begin{theorem}\label{thm:stretchmanifold}
Given a connected curve of a continuous 1-manifold $(\tilde X,\tilde Y)$, there exists a function $\Phi :\mathbb R^2 \rightarrow \mathbb R^2 $ such that  $(\bar X, \bar Y) = \{(\bar x,\bar y)|(\bar x,\bar y)=\Phi(\tilde x,\tilde y)\  \textrm{for}\ \tilde x\in \tilde X, \tilde y\in \tilde Y\}$ is 1-manifold function.  
\end{theorem}

Where $(\bar X,\bar Y)$ is called \textbf{stretched manifold} corresponding to the 1-manifold $(\tilde X,\tilde Y)$ and $\Phi$ is called \textbf{global mirror function}.

\begin{figure}[h!]
\centering
\begin{subfigure}[b]{0.45\textwidth}
  \centering
\includegraphics[height=3.cm]{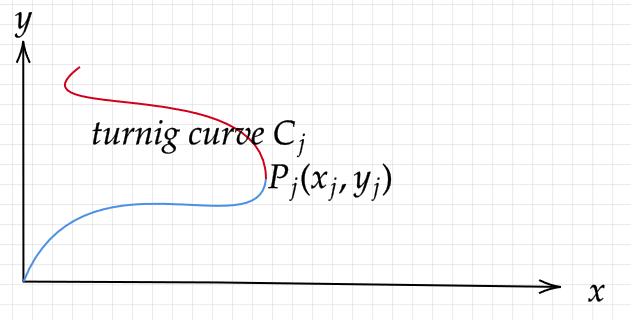}
  \caption{}
  \label{fig:turningcurve_1}
\end{subfigure}%
\\
\begin{subfigure}[b]{0.45\textwidth}
  \centering
\includegraphics[height=3.cm]{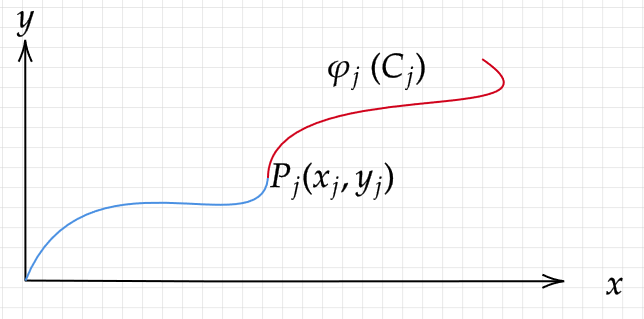}
  \caption{}
  \label{fig:turningcurve_2}
\end{subfigure}%
\caption{(a) Definition of turning point and curve and (b) turning curve is stretched by a mirror function.} \label{fig:turningcurve}
\end{figure}

To elaborate on the theorem~\ref{thm:stretchmanifold}, firstly, we introduce the following concepts: (1) \textbf{turning point}:
    Consider the continuous 1-manifold $(\tilde X, \tilde Y)$. We define turning point $P_j(\tilde x_j,\tilde y_j)$ such that $\Delta \tilde x_j . \Delta \tilde x_{j+1} < 0$ where $\Delta \tilde x_j = \tilde x_j - \tilde x_{j-1}\ \textrm{and}\ \Delta \tilde x_{j+1} = \tilde x_{j+1} - \tilde x_j$ with  $\tilde x_{j-1},\tilde x_j,\tilde x_{j+1}\in \tilde X$, (2) \textbf{turning curve}: 
     $C_j$ (red curve) at turning point $P_j$ shown in Fig.~\ref{fig:turningcurve_1} is a turning curve which is defined by a curve from turning point $P_j$ to another extremity of the 1-manifold, (3) \textbf{mirror function}:
    Next, we define mirror function $\varphi_j:\mathbb R^2 \rightarrow \mathbb R^2$ at a turning point $P_j(\tilde x_j,\tilde y_j)$ such that $\varphi_j(\tilde x,\tilde y) = (2\tilde x_j - \tilde x, \tilde y)$ and Fig.~\ref{fig:turningcurve_2} illustrates the application of the mirror function $\varphi_j$ on the turning curve $C_j$, and (4) \textbf{iterative mirror function}: 
    Consider the initial turning point $P_0$ as one extremity of the curve such that $\tilde y_0 = \min\limits_{\tilde y\in \tilde Y} (\tilde Y)$.  Between the first consecutive turning points $P_0$ and $P_1$, let $\varphi_0(\tilde x,\tilde y)= (\tilde x,\tilde y)$. Furthermore, between $P_j$ and $P_{j+1}$, we apply $j$ times the mirror function $\varphi_i$ at the turning point $P_i$ for $i\in \{0 ... j\}$ then $\Phi_j(\tilde x,\tilde y) = \varphi_j\circ ...\circ\varphi_0(\tilde x,\tilde y)\ \textrm{where}\ \Phi_j=\varphi_j\circ ...\circ\varphi_0$ is called iterative mirror function.

Consider a stretched manifold $(\bar X, \bar Y)$ of the 1-manifold $(\tilde X, \tilde Y)$. Suppose that we have $l$ turning points of the original 1-manifold, $P_i(\tilde x_i,\tilde y_i), i\in \{0...l\}$. $P_0(\tilde x_0,\tilde y_0)$ and $P_{l+1}(\tilde x_{l+1},\tilde y_{l+1})$ are extremities of the 1-manifold. Thus,  corresponding points to the turning points on the stretched manifold is $\bar P_i(\bar x_i,\bar y_i)$ where $(\bar x_i,\bar y_i)=\Phi_i(\tilde x_i,\tilde y_i)$. Therefore, the parametric domain can be written as 
\[\bar X = \bigcup\limits_{i\in \{0...l\}}[\bar x_i,\bar x_{i+1}] = \bigcup\limits_{i\in \{0...l\}}B_i\ \text{and}\ \bar Y = \tilde Y\] 
where $B_i$ is called \textbf{branched parametric domain} (BPD) between turning points $\bar P_{i},\bar P_{i+1}$ associated with the iterative mirror function $\Phi_i$. 

\subsection{Decomposition of parametric domain using reduced order data manifold}

For a given stretched manifold $(\bar X,\bar Y)$, we decompose it into segments w.r.t a defined criterion. This criterion is based on line similarity; Consider three points $\bar P_{j-1}, \bar P_j, \bar P_{j+1}$ on the stretched manifold. \textbf{Line similarity} of the three points is defined such that 
\[
| (\bar P_j - \bar P_{j-1}) \wedge (\bar P_{j+1} - \bar P_{j-1})|<\gamma \frac{\sum\limits_{i\in\{0...n-2\}}|\Delta \bar x_{i+1}|}{\sum\limits_{i\in\{0...n-2\}}|\Delta \bar y_{i+1}|}
\]
where $\gamma$ is a constant which determine the number of the decomposed domain.

The defined threshold allows to keep the average inverse slope of all two consecutive points of the stretched manifold data points to assure that vectors between $\bar P_j, \bar P_{j-1}$ and $\bar P_{j+1}, \bar P_{j-1}$ stay quasi-parallel according to the definition of line similarity. 

At a given point $\bar P_k$, we group the next consecutive points $\bar P_j, j>k$ into the same group of line similarity of the three points $\bar P_k, \bar P_j, \bar P_{j+1}$. If, at $j=k'$, $\bar P_{k'}$ is not in the same group of line similarity $\bar P_k, \bar P_{k'}, \bar P_{k'+1}$, we start to identify the new group of line similarity for the point $\bar P_{k'}$. Besides, if $\bar P_{k+1}$ is not line similarity of the points $\bar P_k,\bar P_{k+1}, \bar P_{k+2}$, we group the two points $\bar P_k, \bar P_{k+1}$ into a group of lines and we start to identify the new group of line similarity of the point $\bar P_{k+2}$. We carry on this procedure till there is no more points on the stretched manifold to be identified.

The pseudo code for the LIne Similarity Segment Decomposition Algorithm (LISSDA) is proposed and is illustrated in the algorithm 2 of the appendix 2.

\subsection{Learning in latent space}

Consider the training data manifold $(X_{train},Y_{train})$ where $X_{train},Y_{train}$ are standardized. For $(x_{train},y_{train})\in (X_{train},Y_{train})$, we find a model $f$ such that $y_{train} = f(x_{train})$. There are many methods in machine learning to learn $f$ by approximating $y_{train}$ as a function of $x_{train}$. The most used one are multilayer perceptron neural networks which can efficiently handle many diverse problems. To be able to enhance its performance, one search to reduce the data dimensionality to learn it in the latent space of the corresponding low dimensional data. We can reduce the high dimensional data manifold to 1-manifold $(\tilde X_{train},\tilde Y_{train})$ where $\tilde x_{train}=\pi_x(x_{train}) \in \tilde X_{train}$ and $\tilde y_{train} = \pi_y(y_{train}) \in \tilde Y_{train}$. Then, we propose to build the stretched manifold $(\bar X_{train},\bar Y_{train})$ from the connected curve of 1-manifold by the global mirror function $\Phi$, then $(\bar x_{train}, \bar y_{train}) = \Phi\circ(\pi_x(x_{train}),\pi_y(y_{train}))$ (theorem~\ref{thm:stretchmanifold}). We can also write $\bar x_{train} = \Phi|_x\circ \pi_x(x_{train})$ and $\bar y_{train} = \Phi|_y\circ \pi_y(y_{train})=\pi_y(y_{train})$ where $\Phi|_x, \Phi|_y$ are the projection of $\Phi$ on $\bar x_{train}$ and $\bar y_{train}$ axes respectively. In the latent space of the stretched manifold $(\bar x_{train},\bar y_{train})\in(\bar X_{train},\bar Y_{train})$, the relation of $\bar x_{train}$ and $\bar y_{train}$ can be written as $\bar y_{train} = g\circ\Phi|_x\circ\pi_x(x_{train})$ where $g$ is any interpolating function of $\bar x_{train}$ and $\bar y_{train}$. To inverse back from $\bar y_{train}$ in the latent space to $y_{train} \in Y_{train}$, we can use the inverse projector $\pi_y^{-1}$. Therefore, we can obtain the approximate $\hat y_{train} \in Y_{train}$ of $y_{train}$ written as $\hat y_{train} = \pi_y^{-1}\circ g\circ\Phi|_x\circ\pi_x(x_{train})$. 

\subsection{Prediction in latent space}

We elaborate in this section a strategy for predicting the output $y_{test}$ for a given $x_{test}$ by the model described in the above section in the latent space. As mentioned in the previous section, consider the original manifold as the manifold of the training data $(X_{train},Y_{train})$, therefore, we can build the global mirror function $\Phi$ and the projectors $\pi_x, \pi_y$ to project the high dimension training data into its corresponding stretched manifold $(\bar X_{train},\bar Y_{train})$.

To predict $y_{test}$ for a given $x_{test}$ according to the training data manifold, firstly, $(x_{test},y_{test})$ is to be identified to which BPD belongs its projection in training data stretched manifold. To do so, the nearest $x_{train}^*$ to the $x_{test}$ is determined by 
$\min\limits_{x_{train}\in X_{train}}d_E(x_{train},x_{test})$.
Secondly, with  $y_{train}^*$ which is the corresponding coordinate of $x_{train}^*$ on the training data manifold, $(x_{train}^*,y_{train}^*)$ is projected into the data training 1-manifold so that its coordinate is $(\tilde x_{train}^*,\tilde y_{train}^*)$. With this coordinate, the corresponding BPD can be identified and so does the corresponding global mirror function $\Phi$. Thus, the projection of $(x_{test},y_{test})$ into the training data stretched manifold can be classified into this BPD. Therefore, we can predict the corresponding output $y_{test}$ in the latent space of stretched manifold by either the inversion of the interpolating function $\hat y_{test} = \pi_y^{-1}\circ g\circ\Phi|_x\circ\pi_x(x_{test})$ or to use directly trained MLPs for each decomposed domain.

\section{RESULTS AND DISCUSSIONS}

\subsection{Harmonic transport problem}\label{sec:problem}

Concerning plane noises, the famous Goldstein equation ~\cite{bensalah2022mathematical} 
is used to model the sound propagation. A simplification of this model is to solve a 1D harmonic transport problem with a high variation source function. The harmonic transport equation is described by the following partial differential equation:
\begin{equation}\label{eq_gsd}
m\frac{\partial y}{\partial x} - i ky(x) - g(x) = 0
\end{equation}

where (1) $i = \sqrt{-1}$, (2) $m$ is mach number, (3) $k$ is wave number, (4) $g$ is the source function, (5)  $x \in [0,1]$  is the spatial domain, and (6) $y(x=0)=0$ is the boundary condition.

The discretized solution of the equation (\ref{eq_gsd}) is $ y \in \mathbb{C}^M$. It can be written as $y_R = (Re(y),Im(y)) \in \mathbb{R}^{2M}$. The source function is presented by a parametric function $g(x) = a e^{\alpha xi} + e^{(\alpha xi - \frac{(x-x_m)^2}{2\sigma^2})}$. The parametric domain is composed of $(m,x_m,k,a,\alpha,\sigma) \in \mathbb{R}^6$. Two cases are considered according to the solution of the equation (\ref{eq_gsd}). For the first case, the output is the scalar defined as the norm of the solution  $||y_R||_{L_2}\in \mathbb{R}$.
, and for the second case, we work on the high dimensional input data where $m=m(x)\in \mathbb{R}^M$ and $g=g(x)\in \mathbb{C}^M$ or $g_R=(Re(g),Im(g))\in \mathbb{R}^{2M}$ are represented by a parametric Chebeshev polynomial that vary at each discretized spatial position with the Chebeshev polynomial coefficient $m\in \mathbb R^M$ and $g_R\in\mathbb R^{2M}$and the parameter $k$ is fixed. So, the input and output data are $p=(m,g_R)\in \mathbb{R}^{3M}$ and $y_R\in \mathbb{R}^{2M}$, respectively.

\subsection{Learning in latent space}\label{sec:app learning latent}

The dataset $(X,Y)$ of the first case is generated by solving the equation~(\ref{eq_gsd}) using Latin Hypercube Sampling (LHS). The parameters' ranges are: $m\in[0.2,0.7]$, $k\in[80.,100.]$, $x_m\in[0.,1.]$, $a\in[0.5,1.]$, $\alpha\in[50.,150.]$, and $\sigma\in[0.1,0.5]$
and as for the second case, the Chebeshev polynimial coefficient of $m$ and $g_R$ are determined by uniform law. The parameters' ranges are: $i\in\{0,...,M-1\}, m_i\in[0.2,0.7]$ and $i\in\{0,...,2M-1\}, g_{R,i}\in[25.,150.]$. The training and test samples are separately generated with the sampling number of $1000$ and $200$, respectively. The two cases of the output described in section~\ref{sec:problem} are considered for the study. 

\begin{figure}[h!]
\centering
\includegraphics[height=6.cm]{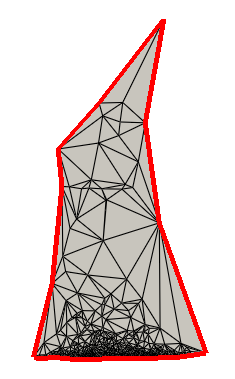}
\caption{Triangulation of 1-manifold by ball pivoting algorithm.} \label{fig:onemanifold_2}
\end{figure}

\begin{figure}[h!]
\centering
\includegraphics[height=8.cm]{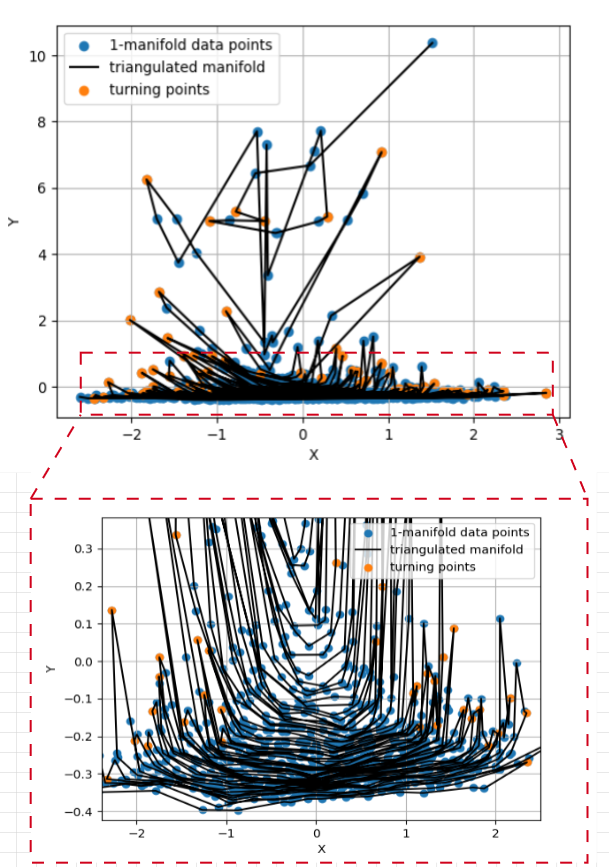}
\caption{Connected curve of the 1-manifold.} \label{fig:connectedcurve}
\end{figure}

\begin{figure}[h!]
\centering
\includegraphics[height=5.cm]{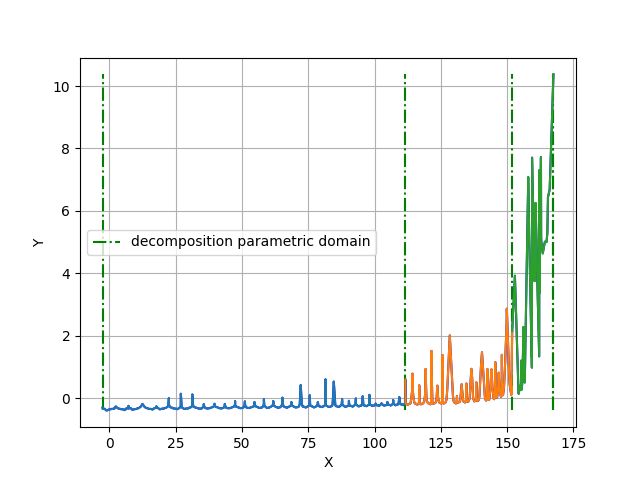}
\caption{Stretched manifold and parametric domain decomposition for the first case with the threshold constant $\gamma=4$.} \label{fig:stretch_2}
\end{figure}

\begin{figure}[h!]
\centering
\includegraphics[height=5.cm]{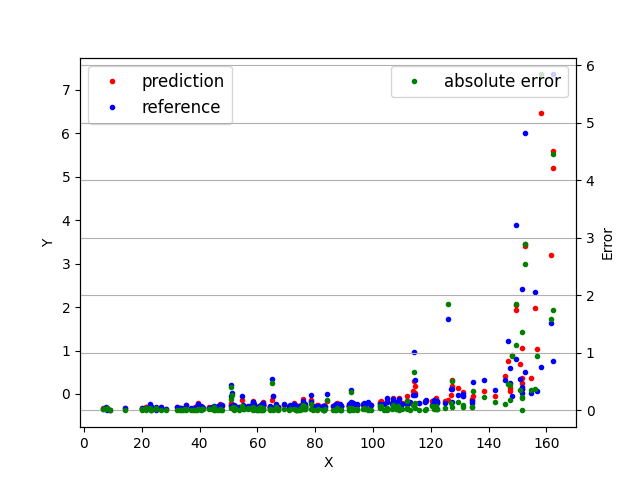}
\caption{References and predictions of the test dataset and their absolute errors.} \label{fig:learning1dinterp}
\end{figure}

We illustrate the methodology for the dataset generated from the first case. For the training data manifold $(X_{train},Y_{train})\subset \mathbb R^6\times \mathbb R$, we use iPCA to reduce $x_{train}\in X_{train}$ to $\tilde x_{train}\in \tilde X_{train}\subset \mathbb R$ where $\tilde x = \pi_x (x)$. Thus, we obtain the 1-manifold $(\tilde X_{train},\tilde Y_{train})$ where $\tilde Y_{train} = Y_{train}$. Next, BPA is used to triangulate the 1-manifold data points as shown in Fig.~\ref{fig:onemanifold_2}. By iteratively extracting the lowest edge of the triangular mesh, we obtain the connected curve shown in Fig.~\ref{fig:connectedcurve}. We observe that the functional continuous higher order manifold is reduced to the spaghetti-like continuous 1-manifold connected curve. After that, we apply the global mirror function to the coordinate points of the 1-manifold connected curve $(\tilde x_{train}, \tilde y_{train})\in (\tilde X_{train}, \tilde Y_{train})$, then we obtain the stretched manifold $(\bar x_{train}, \bar y_{train})=\Phi(\tilde x_{train},\tilde y_{train}) \in (\bar X_{train},\bar Y_{train})$. Then, on the stretched manifold, we use the LISSDA to decompose the parametric domain as illustrated in Fig.~\ref{fig:stretch_2}. As stated before, the number of decomposed domain is depending on the constant $\gamma$. Table ~\ref{tab:gamma_1} illustrates the number of domains according the constant $\gamma$. The larger the constant $\gamma$ is, the larger the average inverse slope is. Hence, the number of domains is smaller. Lastly, in the latent space of the stretched manifold, we can directly predict the output according to each parametric data by applying the interpolating function $g$ on the test dataset $(x_{test},y_{test})\in (X_{test},Y_{test})$, thus, the value of the prediction is obtained via $\hat y_{test} = g\circ\Phi|_x\circ\pi_x(x_{test})$. We depict in the Fig.~\ref{fig:learning1dinterp} the comparison of the predictions by interpolating function and the references in the test dataset and its corresponding absolute errors $|y_{test} - \hat y_{test}|$. We observe that the predictions are very good where the parametric domain of the stretched manifold $\bar x<125$. Clearly, high number of data points is concentrated in the corresponding domain, thus the quality of interpolation is very good. On the other hand, for $\bar x>125$, between each turning points, there are only a few data points. Therefore, the results are not good enough. Furthermore, we can take advantage of the decomposed domain with MLP for the original data manifold, and we compare it with other learning strategies.

\begin{table}[h]
\caption{Number of domains according to the threshold constant $\gamma$ for the first and second case}
\begin{center}
\begin{tabular}{|c|c|c|c|c|c|c|}
\hline
Threshold constant $\gamma$&1&1.5&2&3&4&14\\
\hline
Number of domains&8&7&5&5&3&2\\
\hline
\end{tabular}
\label{tab1}
\end{center}
\label{tab:gamma_1}
\end{table}

\subsection{Benchmark of different learning strategies}

\subsubsection{First case}
In this case, the input data is of dimension $p=6$ and output data of dimension $q=1$. The generated training and testing data described in the previous section for the first case are employed. 

First, the learning of the training data manifold is carried out in the latent space of training data stretched 1-manifold using the classical PCA and the proposed iPCA. The prediction is performed by the interpolating function of both stretched 1-manifolds. 
Table~\ref{tab:comparison_case1} shows the comparison of mean relative error of the two strategies. The error predicted by the learning with iPCA manifold is closed to $100\%$ better than that with PCA manifold. This is directly linked to the data information loss during the construction of the 1-manifold.

Second, as mentioned in section~\ref{sec:app learning latent}, based on the decomposed domain on the stretched 1-manifold using LISSDA, the corresponding data points in the training and testing dataset can be respectively classified into each domain. Different MLP can be used for each domain. The comparison between MLP in full domain and in each decomposed domain is carried out. To be able to evaluate the performance of the MLP in the decomposed domains, we define the weighted error of the decomposed domains as follows:
\[
R_{w} = \frac{\sum\limits_{i\in\{1...\#D\}}N_{test,i}R_{test,i}}{N_{test}},
\]
where $R_{w}$ is the weighted error of all decomposed domain, $N_{test}$ is the number of test samples, $N_{test,i}$ and $R_{test,i}$ are the number of test sample and error, respectively, corresponding to the decomposed domain $D_i\in D$,  and $D$ is the set of all decomposed domains. It should be be noted that each decomposed domain contains at least $100$ samples of the training data points.

Different MLPs for full domain and decomposed domain are tested. A simple architecture is employed for both cases. It is a one layer MLP with 10 nodes and the exponential linear unit $(ELU)$ as activation function. Then, a good learning rate corresponding to full domain and each decomposed domain is chosen. In this case, three decomposed domains are identified for $\gamma=4$. The learning rates $lr = 5.33 10^{-6}, 3.55 10^{-5}, 3.55 10^{-5}, 3.55 10^{-7}$ are corresponding to the full domain, the first, second, and the third decomposed domains, respectively. In all cases, the batchsize and epoch number are $10$ and $2000$ respectively.

Table~\ref{tab:comparison_case1} displays prediction errors and inference time of each strategy. In terms of prediction time, each method performs rapidly and comparably. In terms of precision, for the average global error, the prediction by interpolating function of the stretched 1-manifold with iPCA is the best among all. 
Furthermore, comparing between full domain and decomposed domain MLP, globally, the prediction by the weighted decompose domain provides a better precision. However, in the third domain, the error is very large. It is not only because of the complexity of the manifold in this domain but also because the present of data points are very low, even though, the minimum number of data points of $100$ is imposed. For this case, we didn't use inversion of interpolating function since the output data are scalars. 
We can extend this domain to augment the precision by adding more data either by experiments or by simulations, as well as by statistical methods such as  technique for order preference by similarity \cite{jiang2018adaptive}. However, it is out of scope of the present paper.

\begin{table}[htbp]
\caption{Comparison of different learning models for the first case}
\begin{center}
\begin{tabular}{|c|c|c|c|}
\hline
Relative error&Mean&Variance&t(s)\\
\hline
\hline
\multicolumn{4}{|c|}{interpolating function}\\
\hline
1-Manifold PCA&2.05&3.14&$\sim 10^{-6}$\\
\hline
1-Manifold iPCA&1.04&9.01&$\sim 10^{-6}$\\
\hline
\hline
\multicolumn{4}{|c|}{MLP of original dataset}\\
\hline
Full domain&2.85&11.3&$\sim 10^{-6}$\\
\hline
\hline
First domain&0.09&0.19&$\sim 10^{-6}$\\
\hline
Second domain&1.64&3.96&$\sim 10^{-6}$\\
\hline
Third domain&7.55&15.27&$\sim 10^{-6}$\\
\hline
Weighted domain&1.37&2.98&$\sim 10^{-6}$\\
\hline
\end{tabular}
\label{tab1}
\end{center}
\label{tab:comparison_case1}
\end{table}

\subsubsection{Second case}
In this case, the input and output data are of dimensions $p=3072$ and $q=2048$, respectively. The generated training and testing data described in the section~\ref{sec:app learning latent} are employed. 
First strategy of learning is carried out based on the interpolating function of the stretched manifold. The iPCA is performed on the training dataset to reduce the dimension for the input data. As for the output data, we utilize different dimension reduction methods such as iPCA, classical PCA and UMAP \cite{mcinnes2018umap} to reduce its dimension. The generated 1-manifold of iPCA, PCA and UMAP on training dataset is projected into the corresponding stretched 1-manifold. Thus, the interpolating functions of the training stretched 1-manifold are accordingly established. In order to predict the original output data, the inversion projector of each dimension reduction method is performed on the ordinate of the corresponding stretched 1-manifold.

Table~\ref{tab:comparison_case3} depicts the errors of different learning methods in the latent space by different interpolating functions. Notably, the prediction by UMAP becomes worst among others. It is directly due to insufficient data points of the manifold features that can be captured by UMAP. On the other hand , the prediction by pseudo inverse of iPCA is the best in terms of precision, and also is rapid comparing to others.

Second strategy of the learning is based on MLP of the decomposed domain as described in the first case. The comparison between MLP of the full domain and the decomposed domains is carried out as well.

An architecture of MLP is employed for both cases. It is a one layer MLP with $3087$ nodes and the exponential linear unit $(ELU)$ as activation function. 
In this case, two decomposed domains are identified for $\gamma=2$. The learning rates $lr = 1.4 10^{-3}, 4.72 10^{-5}$ and $4.72 10^{-6}$ are corresponding to the full domain, first, and second decomposed domains, respectively. The batchsizes are $50, 20$ and $20$ corresponding to the full domain, first, and second decomposed domains, respectively. In all cases, epoch number is $2000$ as in the first and second cases.

The findings in Table~\ref{tab:comparison_case3} demonstrate that, when compared to the cases where inversion of interpolated functions are utilized, the MLP for the entire domain provides greater precision. In comparison to the previously mentioned techniques, it is also much faster. Moreover, we observe that the MLP results for the weighted decomposed domains outperform the MLP result for the entire domain by nearly $22\%$.  

\begin{table}[htbp]
\caption{Comparison of different learning models for the third case}
\begin{center}
\begin{tabular}{|c|c|c|c|}
\hline
Relative error&Mean&Variance&t(s)\\
\hline
\hline
\multicolumn{4}{|c|}{Inversion of interpolating function}\\
\hline
Pseudo inverse&1.0008&0.006&0.031\\
\hline
Complement&1.0355&0.63&0.017\\
\hline
PCA&1.0099&0.05&0.021\\
\hline
UMAP&1.1928&0.32&0.133\\
\hline
\hline
\multicolumn{4}{|c|}{MLP of original dataset}\\
\hline
Full domain&0.58&0.30&$\sim 10^{-5}$\\
\hline
\hline
First domain&0.97&0.57&$\sim 10^{-5}$\\
\hline
Second domain&0.36&0.45&$\sim 10^{-5}$\\
\hline
Weighted domain&0.45&0.46&$\sim 10^{-5}$\\
\hline
\end{tabular}
\label{tab1}
\end{center}
\label{tab:comparison_case3}
\end{table}

In addition to the above studies, we include additional study case which is similar to the first case in appendix 3. The inputs of this case are identical to those of the first case, but the outputs are a full solution of the concatenation of the real and image parts of the solution rather than the norm of the solution.

\section{Summary}

We presented a new technique for parametric domain decomposition based on a 1-manifold obtained via a dimensional reduction method. Based on classical principle component analysis (PCA), to keep as much information as possible, we introduced to sequentially reduce the column vectors of the singular vectors of data's singular value decomposition. Then, two approaches for constructing the inverse projector of the iterative were presented. One was based on pseudo inverse of each iteration of the reduced singular vectors, and another was based on the component complement. By use of the generated data for harmonic transport equation, along with the assumption of the continuous manifold function of the original data, the methodology of constructing the stretched 1-manifold was presented.
On top of that, an algorithm based on line similarity (LISSDA) was proposed to decompose the parametric domain of the stretched 1-manifold. 
Moreover, to demonstrate the performance of the proposed technique, the benchmark on two proposed studied cases were carried out on different learning methods. These methods were classified into two strategies : inversion of interpolating function in the latent space and MLPs for the decomposed parametric domains using the original data. For a low dimensional output, the interpolating function showed a very good performance comparing to MLP. Nevertheless, for a high dimensional output, MLP for the decomposed domains globally exhibits a very good performances comparing to other methods. Moreover, the MLPs were the least computationally expensive methods for prediction. 

The proposed technique gives an insight how to exploit the properties of the data manifold in a lower dimension, however, this methodology should be tested for other cases, especially those who exhibit nonlinear behavior.

\section*{Acknowledgements}
The authors also thank the industrial partners for their supports in the framework of the HSA project \cite{irtsysx}. This project is as well supported by the French government's aid in the framework of PIA (Programme d'Investissement d'Avenir) for Institut de Recherche Technologique SystemX.
\clearpage
\bibliographystyle{unsrt} 
\bibliography{references}

\clearpage

\section*{APPENDIX 1: PROOF OF THEOREMS}

\begin{proof}[Proof of Theorem 2.1]
Given that the original data $X=X_0$ if of dimension $p\in \mathbb N$, we use SVD of the original data $X_0= U_0 S_0 V_0$. Thus, the reduced components can be written as $X_1=X_0 V_1$ where $V_1$ are the reduced singular vectors of $V_0$ by eliminating its column vectors according to either $r\%$ or $s$. By fixing $r\%$ as a satisfied amount of information at each iteration $j$ of the reduced component, the dimension of $X_j$ is $p_j>p'$. If at iteration $j'$, the full components are needed, then we reduced by $s=1$ which is the last component of singular vector $V_{j'}$. By iteratively doing so, we can reduce $p$ to $p'$ with a finite number of iteration, assuming it to be $k$. Thus, $X_k=X_{k-1}V_k=...=X_0V_1...V_k=X_0\pi$.
\end{proof}

\begin{proof}[Proof of Theorem 2.2]
Given the original data $X_0$, at first iteration, $X_0=X_1F_1=X_2F_2F_1=...=X_kF_k...F_1=X_k\pi_k^{-1}$
\end{proof}

\begin{proof}[Proof of Theorem 2.3]
At iteration $j$, because $V_j$ and $V_j^c$ are complement to each other, we can write $V_{j-1}=[V_j|V_j^c]$. Therefore, $\mathbb I = V_{j-1}V_{j-1}^T = [V_j|V_j^c][V_j|V_j^c]^T=V_jV_j^T+V_j^cV_j^{cT}$. 

We can write : $X_jV_j^T=X_{j-1}V_jV_j^T$ and $X_j^cV_j^{cT}=X_{j-1}V_j^cV_j^{cT}$. Thus, $X_{j-1}=X_jV_j^T+X_j^cV_j^{cT}$.

As $X'_k$ is known in $X'_{k-1}=X'_kV_j^T+X_k^{c'}V_k^{cT}$, we need to find $X_k^{c'}$. If $X'_k\in X_k$, then either $X_k^{c'}\in X_k^c$ or it is the solution to the minimize problem $\min\limits_{X_k^{c'}}d_E((X_k',X_k^{c'}),(X_k,X_k^c))$. Then, we obtain $X'_{k-1}$. By iteratively applying the same process, we can obtain $X_0$.

\end{proof}

\begin{proof}[Proof of Theorem 2.4]
Given a connected curve of 1-manifold $(\tilde X,\tilde Y)$. At the first turning point $P_1$, by applying the mirror function $\varphi_1$ on the turning curve $C_1$,  the curve defined from the initial turning point $P_0$ to $P_1$ has a unique image corresponding to each elements of the domain of this part of the curve. As for the curve from $P_1$ to another extremity, we determine the next turning point $P_2$. Suppose that, at turning point $P_k$ and its corresponding turning curve $C_k$, the curve from the initial turning point to the turning point $P_i$, for $i\in\{0...k\}$, which is applied by the mirror function $\Phi_i = \varphi_i \circ \varphi_{i-1} ...\circ \varphi_0$, is a 1-manifold function of this part of the curve. Then, on the turning curve $\Phi_k(C_k)$, we determine the next turning point $P_{k+1}$ and its corresponding turning curve $C_{k+1}$. By applying the mirror function $\Phi_{k+1}$ on $C_{k+1}$, the part of the curve from $P_k$ to $P_{k+1}$ becomes a functional curve. Therefore, there exists a function $\Phi$ such that $\Phi(C_0)$ is 1-manifold function where $C_0$ is the initial turning curve or initial curve. The function is a defined as $\Phi:\mathbb R^2 \rightarrow \mathbb R^2$ and $\Phi(P) = \Phi_i(P)\ \textrm{if}\ P\in [P_i,P_{i+1}]\ \textrm{of the turning curve}\ C_i$.
\end{proof}
\clearpage

\section*{APPENDIX 2: DECOMPOSITION ALGORITHMS}

\begin{algorithm}[h!]
\caption{LIne Similarity Points (LISP)}\label{algo:lsp}
\begin{algorithmic}
  \State $C \gets$ stretched manifold points
  \Function{LISP}{C}
    \State $L \gets$ empty list of points on a line
    \State ok $\gets$ True
    \State $n,i \gets \#C - 1,1$
    \State $p_1 \gets C[0]$
    \While{$\#C > 0$ and $i < n$}
        \State $i \gets i + 1$
        \State $p_2,p_3 \gets C[i - 1],C[i]$
        \If{$|(p_2 - p_1)\times (p_1 - p_3)| > \epsilon$}
            \State ok $\gets$ False
            \If{$i > 2$}
                \State $L \gets C[:i]$
                \State $C \gets C[i-1:]$
                \State $n,i \gets \#C - 1,1$
            \EndIf
            \State break
        \EndIf
    \EndWhile
    
   \Return $L,C,ok$
   \EndFunction
\end{algorithmic}
\end{algorithm}

\begin{algorithm}[h!]
\caption{LISSDA}\label{algo:lssd}
\begin{algorithmic}
  \State $C \gets$ stretched manifold points
  \Function{LISSDA}{C}
     \State $L_{total} \gets$ empty list of decomposed line
     \While{$\#C > 0$}
        \State $L,C,ok \gets$ LISP$(C)$
        \If{$ok==True$}
            \State $L \gets C$
            \State $C \gets$ empty list
        \Else
            \State $L' \gets C[:2]$
            \State $C \gets C[1:]$
        \EndIf
        \State $L_{total} \gets L_{total} + [L,L']$
    \EndWhile
    
   \Return $L_{total}$
   \EndFunction
\end{algorithmic}
\end{algorithm}

\clearpage 

\section*{APPENDIX 3: ADDITIONAL STUDY CASE}

In this case, the input data is of dimension $p=6$ and output data of dimension $q=2048$. First strategy of learning is carried out in the same way as described in the second case. 
Table~\ref{tab:comparison_case2} shows the error of the vector output based on norm $L_{\infty}$ averaging on total test samples. The prediction by inverse UMAP is the best among other methods. UMAP is good when the stretched manifold is able to capture the original manifold features. It allows UMAP to preserve most information during the dimension reduction for inverting to the original data. However, this method is the most computationally expensive. On the other hand, the prediction by the pseudo inverse of iPCA is the second best (among different interpolating functions) and its inference time is also second best comparing to PCA. iPCA could be better than PCA, only when, at its each iteration, EVR is kept at its maximum before reaching one dimension. The method of component complement of iPCA seems to be less efficient among others. The drawback of this method is linked to the computation of the interpolation of the complement coordinate. The method can give more precise prediction only if the data manifold points are closed to each other. In our case, the data points are scarce comparing to the data manifold dimension.

Second strategy of the learning is based on use of MLPs for the decomposed domains as described in the first case. The comparison between MLP for the full domain and the decomposed domains is carried out as well.

An architecture of MLP is employed for both cases. It is a two-layer MLP with $1345$ and $3680$ nodes respectively and the exponential linear unit $(ELU)$ as activation function. Then, a good learning rate corresponding to full domain and each decomposed domain is chosen. In this case, three decomposed domains are identified for $\gamma=1.5$. The learning rates $lr = 1.11 10^{-7}, 1.24 10^{-7}, 1.24 10^{-6}$ and $6.2 10^{-9}$ are corresponding to the full domain, first, second and third decomposed domains, respectively. The batchsizes are $50, 30, 30$ and $5$ corresponding to the full domain, first, second and third decomposed domains, respectively. In all cases, epoch number is $2000$ as in the first case.

The results in Table~\ref{tab:comparison_case2} show that the MLP for the full domain offers better precision compared to the cases where inversion of interpolated functions are used. It is also considerably faster than the aforementioned methods. Furthermore, we can see that the results of MLPs for the decomposed domains are almost $10\%$ better than the result of the MLP for the full domain.  

\begin{table}[htbp]
\caption{Comparison of different learning models for the second case}
\begin{center}
\begin{tabular}{|c|c|c|c|}
\hline
Error&Mean&Variance&t(s)\\
\hline
\hline
\multicolumn{4}{|c|}{Inversion of interpolating function}\\
\hline
pseudo inverse&1.0047&0.012&0.034\\
\hline
Complement &1.5296&1.06&0.084\\
\hline
PCA&1.0196&0.053&0.023\\
\hline
UMAP&1.0008&0.047&0.139\\
\hline
\hline
\multicolumn{4}{|c|}{MLP of original dataset}\\
\hline
Full domain&0.8566&0.12&$\sim 10^{-5}$\\
\hline
\hline
First domain&0.8125&0.18&$\sim 10^{-5}$\\
\hline
Second domain&0.7079&0.36&$\sim 10^{-5}$\\
\hline
Third domain&1.0654&0.15&$\sim 10^{-5}$\\
\hline
Weighted domain&0.7779&0.27&$\sim 10^{-5}$\\
\hline
\end{tabular}
\label{tab1}
\end{center}
\label{tab:comparison_case2}
\end{table}

\end{document}